\documentclass{article} 
\usepackage{iclr2015,times}
\usepackage{url}
\usepackage{multirow}
\usepackage{float}
\usepackage[caption = false]{subfig}
\usepackage{graphicx}

\usepackage{amsmath}
\usepackage{amsfonts}
\usepackage{bm}
\newcommand{\mat}[1]{\boldsymbol{\mathbf{#1}}}
\newcommand{\argmin}{\operatornamewithlimits{arg\ min}}

\newcommand{\bb}[1]{\mathbb{#1}}
\newcommand{\reg}[1]{\mathrm{#1}}

\title{Improving zero-shot learning by mitigating the hubness problem}

\author{
Georgiana Dinu, Angeliki Lazaridou, Marco Baroni\\
Center for Mind/Brain Sciences\\
University of Trento (Italy)\\
\texttt{georgiana.dinu|angeliki.lazaridou|marco.baroni@unitn.it} \\
}

\iclrfinalcopy 

\begin{document}

\maketitle

\begin{abstract}
  The zero-shot paradigm exploits vector-based word representations
  extracted from text corpora with unsupervised methods to learn
  general mapping functions from other feature spaces onto word space,
  where the words associated to the nearest neighbours of the mapped
  vectors are used as their linguistic labels. We show that the
  neighbourhoods of the mapped elements are strongly polluted by
  \emph{hubs}, vectors that tend to be near a high proportion of
  items, pushing their correct labels down the neighbour list. After
  illustrating the problem empirically, we propose a simple method to
  correct it by taking the proximity distribution of potential
  neighbours across many mapped vectors into account. We show that
  this correction leads to consistent improvements in realistic
  zero-shot experiments in the cross-lingual, image labeling and image
  retrieval domains.\end{abstract}

\section{Introduction}
\label{sec:introduction}

Extensive research in computational linguistics and neural language
modeling has shown that contextual co-occurrence patterns of words in
corpora can be effectively exploited to learn high-quality
vector-based representations of their meaning in an unsupervised
manner
\citep{Collobert:etal:2011,Clark:2012b,Turney:Pantel:2010}. This
has in turn led to the development of the so-called \emph{zero-shot
  learning} paradigm as a way to address the manual annotation
bottleneck in domains where other vector-based representations (e.g.,
images or brain signals) must be associated to word labels
\citep{Palatucci:etal:2009}. The idea is to use the limited training
data available to learn a general mapping function from vectors in the
domain of interest to word vectors, and then apply the induced
function to map vectors representing new entities (that were not seen
in training) onto word space, retrieving the nearest neighbour words
as their labels. This approach has originally been tested in neural
decoding \citep{Mitchell:etal:2008,Palatucci:etal:2009}, where the
task consists in learning a regression function from fMRI activation
vectors to word representations, and then applying it to the brain
signal of a concept outside the training set, in order to ``read the
mind'' of subjects. In computer vision, zero-shot mapping of image
vectors onto word space has been applied to the task of retrieving
words to label images of objects outside the training inventory
\citep{Frome:etal:2013,Socher:etal:2013a}, as well as using the inverse
language-to-vision mapping for image retrieval
\citep{Lazaridou:etal:2014}. Finally, the same approach has been
applied in a multilingual context, using translation pair vectors to
learn a cross-language mapping, that is then exploited to translate
new words \citep{mikolov2013exploiting}.

Zero-shot learning is a very promising and general technique to reduce
manual supervision. However, while all experiments above report very
encouraging results, performance is generally quite low in absolute
terms. For example, the system of \cite{Frome:etal:2013} returns the
correct image label as top hit in less than 1\% of cases in all
zero-shot experiments (see their Table 2). Performance is always above
chance, but clearly not of practical utility.

In this paper, we study one specific problem affecting the quality of
zero-shot labeling, following up on an observation that we made,
qualitatively, in our experiments: The neighbourhoods surrounding
mapped vectors contain many items that are ``universal'' neighbours,
that is, they are neighbours of a large number of different mapped
vectors. The presence of such vectors, known as \emph{hubs}, is an
intrinsic problem of high-dimensional spaces
\citep{Radovanovic:etal:2010b}. Hubness has already been shown to be
an issue for word-based vectors
\citep{Radovanovic:etal:2010a}.\footnote{\cite{Radovanovic:etal:2010a}
  propose a supervised hubness-reducing method for document vectors
  that is not extensible to the zero-shot scenario, as it assumes a
  binary relevance classification setup.} However, as we show in Section 2, the problem is much
more severe for neighbourhoods of vectors that are mapped onto a
high-dimensional space from elsewhere through a regression
algorithm. We leave a theoretical understanding of \emph{why} hubness affects
regression-based mappings to further work. Our current contributions
are to demonstrate the hubness problem in the zero-shot setup, to
present a simple and efficient method to get rid of it by adjusting
the similarity matrix after mapping, and to show how this brings
consistent performance improvements across different tasks. While one
could address the problem by directly designing hubness-repellent
mapping functions, we find our post-processing solution more
attractive as it allows us to use very simple and general
least-squares regression methods to train and perform the mapping.

We use use the term \emph{pivots} to stand for a set of vectors we
retrieve neighbours for (these comprise at least, in our setting, the
zero-shot-mapped vectors) and \emph{targets} for the subspace of
vectors we retrieve the neighbours from (often, corresponding to the
whole space of interest). Then, we can phrase our proposal as
follows. Standard nearest neighbour queries rank the targets
independently for each pivot. A single target is allowed to be the
nearest neighbour, or among the top $k$ nearest neighbours, of a large
proportion of pivots: and this is exactly what happens empirically
(the hubness problem). We can greatly mitigate the problem by taking
the \emph{global distribution of targets across pivots} into
account. In particular, we use the very straightforward and effective
strategy of inverting the query: we convert the similarity scores of a
\emph{target} with all pivots to the corresponding ranks, and then
retrieve the nearest neighbours of a pivot based on such ranks,
instead of the original similarity scores. We will empirically show
that with this method high-hubness targets are down-ranked for many
pivots, and will kept as neighbours only when semantically
appropriate.

\section{Hubness in zero-shot mapping}
\label{sec:hubness-problem}

\paragraph{The Zero-shot setup}
In zero-shot learning, training data consist of vector representations
in the \emph{source} domain (e.g., source language for translation,
image vectors for image annotation) paired with language labels (the
\emph{target} domain): $D_{tr} = \{(\mat{x}_i,y_i)\}_{i=1}^{m}$, where
$\mat{x}_i\in \bb{R}^u$ and $y_i\in T_{tr}$, a vocabulary containing
training labels. At test time, the task is to label vectors which have
a novel label: $D_{ts} = \{(\mat{x}_i,y_i)\}_{i=1}^{n}$, $y_i\in
T_{ts}$, with $T_{ts} \cap T_{tr} = \emptyset$. This is possible
because labels $y$ have vector representations $\mat{y} \in
\mathbb{R}^v$.\footnote{We use $x$ and $\mat{x}$ to stand for a label
  and its corresponding vector.}  Training is cast as a multivariate
regression problem, learning a function which maps the source domain
vectors to their corresponding target (linguistic-space) vectors.  A
straightforward and performant choice
\citep{Lazaridou:etal:2014,mikolov2013exploiting} is to assume the
mapping function is a linear map $\mat{W}$, and use a l2-regularized
least-squares error objective:
\begin{equation}
\hat{\mat{W}} = \argmin_{\mat{W} \in \bb{R}^{v \times u}}||\mat{X}\mat{W}-\mat{Y}||_F + \lambda||\mat{W}||
\label{eq:obj1}
\end{equation}
where \mat{X} and \mat{Y} are matrices obtained through the
concatenation of train source vectors and the target vectors of the
corresponding labels.

Once the linear function has been estimated, any source vector
$\mat{x}\in\bb{R}^u$ can be mapped into the target domain through
$\mat{x}^T\mat{W}$.

\paragraph{Target space label retrieval}
Given a source element $x \in S$ and its vector $\mat{x}$, the
standard way to retrieve a target space label ($T$) is by returning
the nearest neighbour (according to some similarity measure) of mapped
$\mat{x}$ from the set of vector representations of $T$. Following
common practice, we use the cosine as our similarity measure.

We denote by $\reg{Rank}_{x,T}(y)$ the rank of an element $y \in T$
w.r.t.~its similarity to $x$ and assuming a query space $T$. More
precisely, this is the position of $y$ in the (decreasingly) sorted
list of similarities: $[cos(x,y_i)|y_i \in T]$. This is an integer
from $1$ to $|T|$ (assuming distinct cosine values).  Under this
notation, the standard nearest neighbour of $x$ is given by:
\begin{equation}
\reg{NN}_1(x,T) = \argmin_{y\in T}\reg{Rank}_{x,T}(y)
\label{baseline}
\end{equation}
We will use $\reg{NN}_k(x,T)$ to stand for the set of $k$-nearest
neighbours in $T$, omitting the $T$ argument for brevity.

\paragraph{Hubness} We can measure how \emph{hubby} an item $y \in T$
is with respect to a set of pivot vectors $P$ (where $T$ is the search
space) by counting the number of times it occurs in the $k$-nearest
neighbour lists of elements in $P$:
\begin{equation}
\reg{N}_{k,P}(y) = |\{y \in \reg{NN}_k(x,T)|x \in P\}|
\label{hub-def}
\end{equation}
An item with a large $\reg{N}_{k}$ value (we will omit the set
subscript when it is clear from the context) occurs in the
$\reg{NN}_k$ set of many elements and is therefore a hub.

Hubness has been shown to be an intrinsic problem of high-dimensional
spaces: as we increase the dimensionality of the space, a number of
elements, which are, by all means, \emph{not} similar to all other
items, become hubs. As a results nearest neighbour queries return the
hubs at top 1, harming accuracy. It is known that the problem of
hubness is related to \emph{concentration}, the tendency of pairwise
similarities between elements in a set to converge to a constant as
the dimensionality of the space increases
\citep{Radovanovic:etal:2010b}. \cite{Radovanovic:etal:2010a} show
that this also holds for cosine similarity (which is used almost
exclusively in linguistic applications): the expectation of pairwise
similarities becomes constant and the standard deviation converges to
0. This, in turn, is known to cause an increase in hubness.

\paragraph{Original vs.~mapped vectors} In previous work
we have (qualitatively) observed a tendency of the hubness problem to
become worse when we query a target space in which some elements have
been mapped from a different source space. In order to investigate
this more closely, we compare the properties of mapped elements versus
original ones. We consider word translation as an application and use
300-dimensional vectors of English words as source and vectors of
Italian words as target. We have, in total, vocabularies of 200,000
English and Italian words, which we denote $S$ and $T$. We use a
set of 5,000 translation pairs as training data and learn a linear map.

\begin{figure}
\centering
\subfloat[Original]{\includegraphics[scale=0.40]{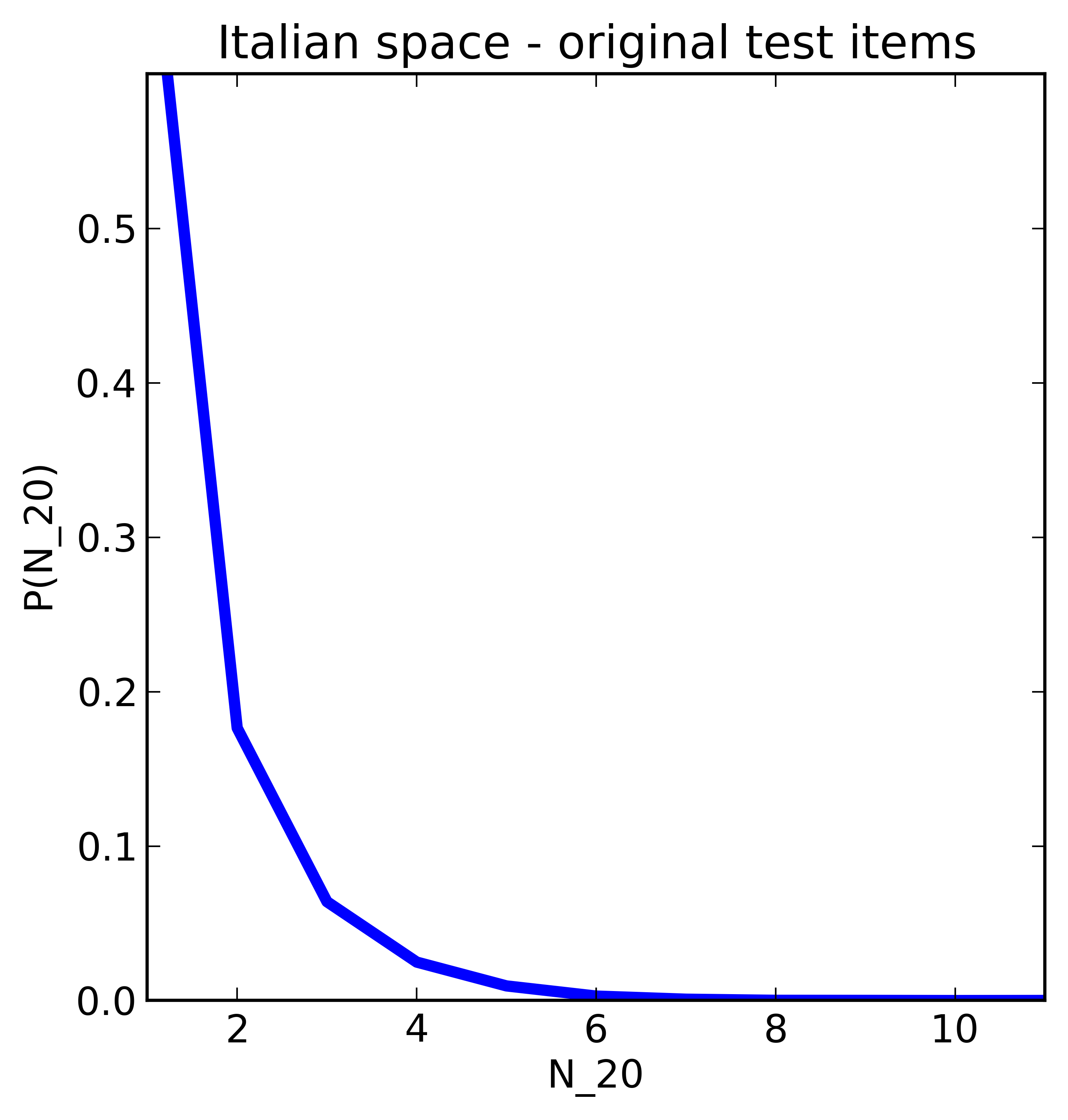}} 
\subfloat[Mapped, with regularization]{\includegraphics[scale=0.40]{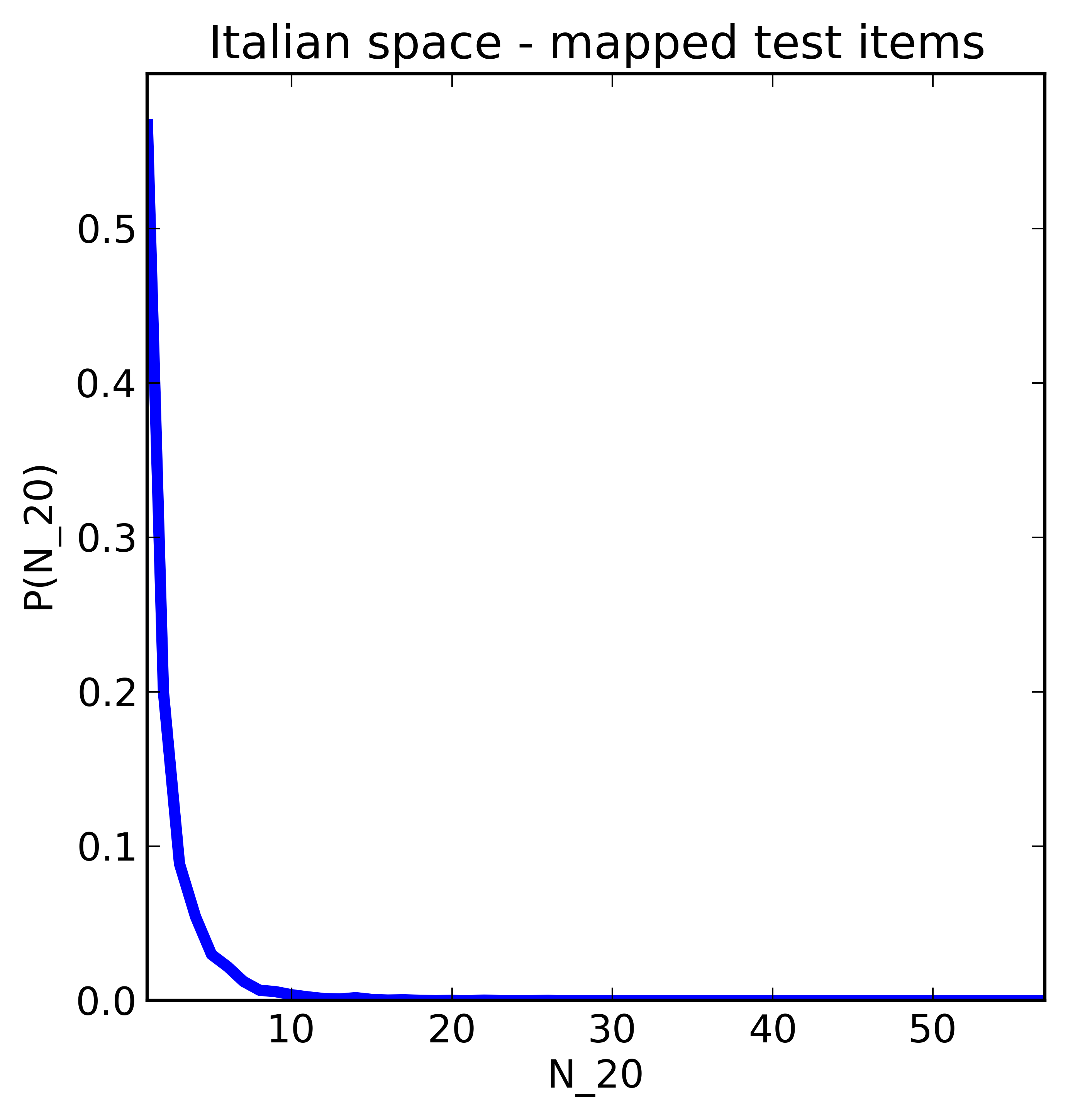}} 
\subfloat[Mapped, no regularization]{\includegraphics[scale=0.40]
{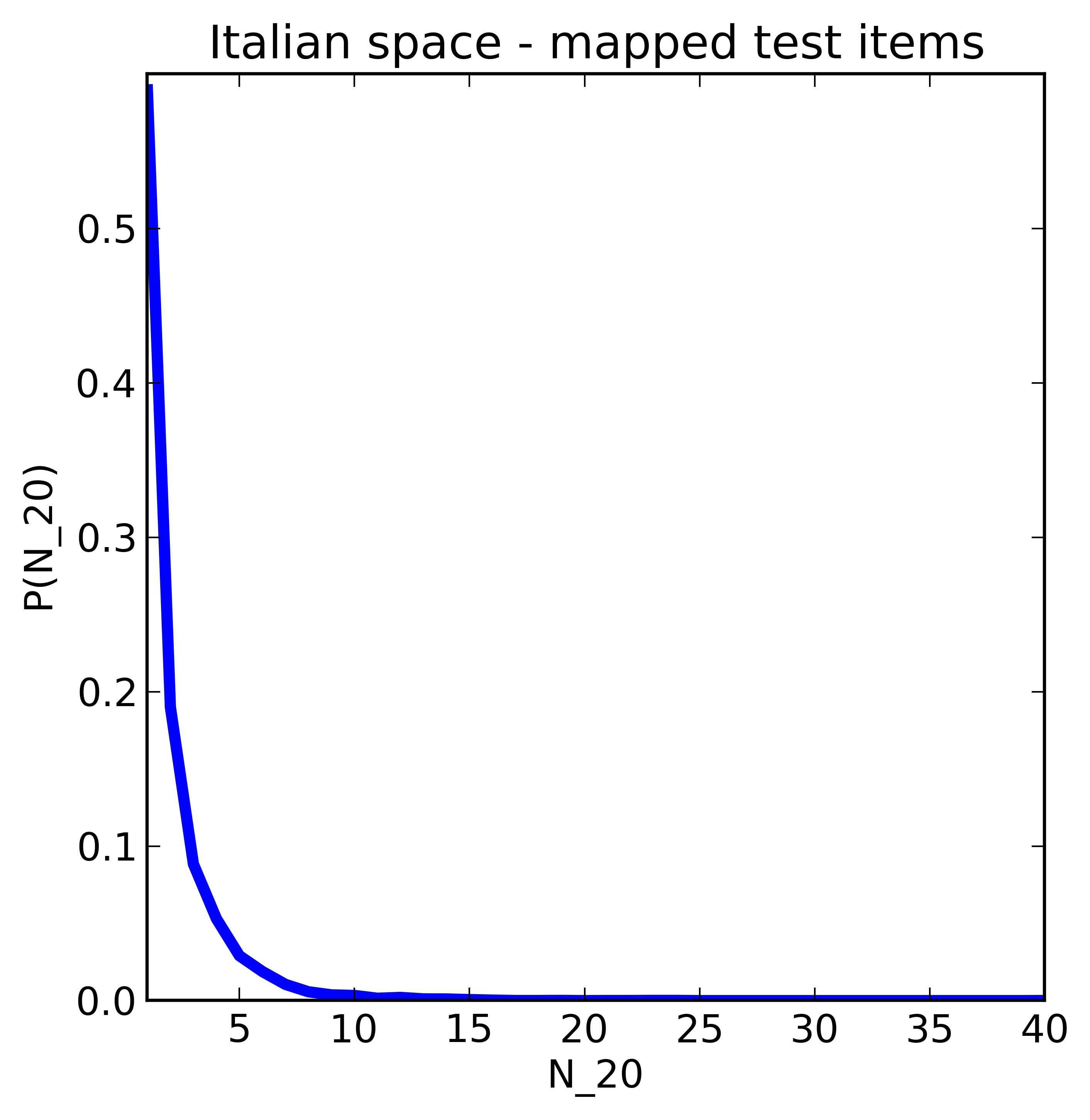}}
\caption{Distribution of $\reg{N}_{20}$ values of target space
  elements ($\reg{N}_{20,Ts}(y)$ for $y\in T$). Test elements (pivots)
  in the target space (original) vs.~corresponding vectors obtained by
  mapping from the source space (mapped).  Significantly larger
  $\reg{N}_{20}$ values are observed in the mapped spaces (maxima at
  57 and 40 vs.~11 in original space.)}
\label{problem-language-1}
\end{figure}

We then pick a random test set $Ts$ of 1,500 English words that have
not been seen in training and map them to Italian using the learned
training function (full details in Section
\ref{sec:english-to-italian} below). We compute the hubness of all
elements in $T$ using the test set items as pivots, and considering
all $200,000$ items in the target space as potential neighbours (as
any of them could be the right translation of a test word). In the
first setting (\emph{original}), we use target space items: for the
test instance $car\to auto$, we use the true Italian $auto$ vector. In
the second and third settings (\emph{mapped}) we use the mapped
vectors (our predicted translation vector of $car$ into Italian),
mapped through a matrix learned without and with regularization,
respectively. Figure \ref{problem-language-1} plots the distribution
of the $\reg{N}_{20,Ts}(y)$ scores in these three settings.

As the plots show, the hubness problem is indeed greatly
exacerbated. When using the original $Ts$ elements, target space hubs
reach a $\reg{N}_{20}$ level of at most 11, meaning they occur in the
$\reg{NN}_{20}$ sets of 11 test elements. On the other hand, when
using mapped elements the maximum $\reg{N}_{20}$ values are above 40
(note that the $x$ axes are on different scales in the
plots!). Moreover, regularization does not significantly mitigate
hubness, suggesting that it is not just a matter of overfitting, such
that the mapping function projects everything near vectors it sees
during training.

\section{A globally corrected neighbour retrieval method}

One way to correct for the increase in hubness caused by mapping is to
compute hubness scores for all target space elements. Then, given a
test set item, we re-rank its nearest neighbours by downplaying the
importance of elements that have a high hubness score. Methods for
this have been proposed and evaluated, for example, by
\cite{Radovanovic:etal:2010a} and \cite{10.1109/ICCP.2011.6047899}. We
adopt a much simpler approach (similar in spirit to
\citealp{conf/cikm/TomasevRMI11}, but greatly simplified), which takes
advantage of the fact that we almost always have access not to just 1
test instance, but more vectors in the source domain (these do not
need to be labeled instances). We map these additional pivot elements
and conjecture that we can use the topology of the subspace where the
mapped pivot set lives to correct nearest neighbour retrieval. We
consider first the most straightforward way to achieve this effect. A
hub is an element which appears in many $\reg{NN}_k$ lists because it
has high similarity with many items. A simple way to correct for this
is to normalize the vector of similarities of each target item to the
mapped pivots to length 1, prior to performing $\reg{NN}$ queries.
This way, a vector with very high similarities to many pivots will be
penalized. We denote this method $\reg{NN}_\reg{nrm}$.

We propose a second corrected measure, which does not re-weight the
similarity scores, but ranks target elements using $\reg{NN}$
statistics for the entire mapped pivot set. Instead of the nearest
neighbour retrieval method in Equation (\ref{baseline}), we use a
following \emph{globally-corrected} ($\reg{GC}$) approach, that could
be straightforwardly implemented as:
\begin{equation}
\reg{GC}_1(x,T) = \argmin_{y\in T}\reg{Rank}_{y,P}(x)
\label{eq:corrected-first-take}
\end{equation}
To put it simply, this method reverses the querying: instead of
returning the nearest neighbour of pivot $x$ as a solution, it returns
the target element $y$ which has $x$ ranked highest. Intuitively, a
hub may still occur in the $\reg{NN}$ lists of some elements, but only
if not better alternatives are present. The formulation of GC in
Equation (\ref{eq:corrected-first-take}) can however lead to many tied
ranks: For example, we want to translate $car$, but both Italian
$auto$ and $macchina$ have $car$ as their second nearest neighbour (so
both rank 2) and no Italian word has $car$ as first neighbour (no rank
1 value). We use the cosine scores to break ties, therefore $car$ will
be translated with $auto$ if the latter has a higher cosine with the
mapped $car$ vector, with $macchina$ otherwise. Note that when only
one source vector is available, the GC method becomes equivalent to a
standard $\reg{NN}$ query. As the cosine is smaller than 1 and ranks
larger or equal to 1, the following equation implements GC with
cosine-based tie breaking:
\begin{equation}
\reg{GC}_1(x,T) = \argmin_{y\in T}(\reg{Rank}_{y,P}(x)-cos(x,y))
\label{corrected}
\end{equation}

\subsection{English to Italian word translation}
\label{sec:english-to-italian}

We first test our methods on bilingual lexicon induction. As the
amount of parallel data is limited, there has been a lot of work on
acquiring translation dictionaries by using vector-space methods on
monolingual corpora, together with a small seed lexicon
\citep{HaghighiLBK08,Klementiev:2012,Koehn02learninga,Rapp99}. One
of the most straightforward and effective methods is to represent
words as high-dimensional vectors that encode co-occurrence only with
the words in the seed lexicon and are therefore comparable
cross-lingually \citep{Klementiev:2012,Rapp99}. However, this method
is limited to vector spaces that use words as context features, and
does not extend to vector-based word representations relying on other
kinds of dimensions, such as those neural language models that have
recently been shown to greatly outperform context-word-based
representations \citep{Baroni:etal:2014}. The zero-shot approach, that
induces a function from one space to the other based on paired seed
element vectors, and then applies it to new data, works irrespective
of the choice of vector representation. This method has been shown to
be effective for bilingual lexicon construction by
\cite{mikolov2013exploiting}, with \cite{Dinu:Baroni:2014} reporting
overall better performance than with the seed-word-dimension
method. We set up a similar evaluation on the task of finding Italian
translations of English words.

\paragraph{Word representations}
The \emph{cbow} method introduced by \cite{Mikolov:etal:2013b} induces
vector-based word representations by trying to predict a target word
from the words surrounding it within a neural network architecture. We
use the word2vec
toolkit\footnote{\url{https://code.google.com/p/word2vec/}} to learn
300-dimensional representations of 200,000 words with cbow. We
consider a context window of 5 words to either side of the target, we
set the sub-sampling option to 1e-05 and estimate the probability of a
target word with the negative sampling method, drawing 10 samples from
the noise distribution \citep{Mikolov:etal:2013b}. We use 2.8 billion
tokens as input (ukWaC + Wikipedia + BNC) for English and the 1.6
billion itWaC tokens for Italian.\footnote{Corpus sources:
  \url{http://wacky.sslmit.unibo.it}, \url{http://en.wikipedia.org},
  \url{http://www.natcorp.ox.ac.uk}}

\paragraph{Training and testing} Both train and test translation pairs
are extracted from a dictionary built from Europarl, available at
\url{http://opus.lingfil.uu.se/} (Europarl, en-it)
\citep{TIEDEMANN12.463}. We use 1,500 English words split into 5
frequency bins as test set (300 randomly chosen in each bin). The bins
are defined in terms of rank in the (frequency-sorted) lexicon:
[1-5K], [5K-20K], [20K-50K], [50K-100K] and [100K-200K]. The bilingual
lexicon acquisition literature generally tests on very frequent words
only. Translating medium or low frequency words is however both more
challenging and useful. We also sample the training translation pairs
by frequency, using the top $1K$, $5K$, $10K$ and $20K$ most frequent
translation pairs from our dictionary (by English frequency), while
making sure there is no overlap with test elements.

For each test element we query the entire (200,000) target space and
report translation accuracies.  An English word may occur with more
than one Italian translation (1.2 on average in the entire data): in
evaluation, an instance is considered correct if any of these is
predicted.  We test the standard method (regular $\reg{NN}$ querying)
as well as the two corrected methods: $\reg{NN}_{\reg{nrm}}$ and
$\reg{GC}$. As previously discussed, the latter benefit from more
mapped data, in addition to an individual test instance, to be used as
pivots. In addition to the 1,500 test elements, we report performance
when mapping other 20,000 randomly chosen English words (their Italian
translations are not needed). We actually observed improvements also
when using solely the 1,500 mapped test elements as pivots, but
increasing the size with arbitrary additional data (that can simply be
sampled from the source space without any need for supervision) helps
performance.

\begin{figure}
\begin{minipage}[c]{0.48\linewidth}
\begin{tabular}{c|ccc}
  Train size & \multicolumn{3}{|c}{No regularization}\\ 
  \hline
  & NN & NN$_{\reg{nrm}}$ & GC \\
  1K  & 14.9 & 20.9 & 20.9        \\ 
  5K  & 30.3 & 33.1 & \textbf{37.7}\\ 
  10K & 30.0 & 33.5 & \textbf{37.5}\\ 
  20K & 25.1 & 32.9 & \textbf{37.9}\\ 
\end{tabular}
\vskip 0.3cm
\begin{tabular}{c|ccc}
Train size & \multicolumn{3}{|c}{GCV}\\ 
\hline
		& NN & NN$_{\reg{nrm}}$ & GC \\
1K  & 12.4 & 25.5 & \textbf{28.7}\\
5K  & 27.7 & 32.9 & 37.5 \\
10K & 28.2 & 33.1 & 37.3\\
20K & 23.7 & 32.0 & 37.8
\end{tabular}
\end{minipage}
\begin{minipage}[c]{0.48\linewidth}
\includegraphics[scale=0.50]{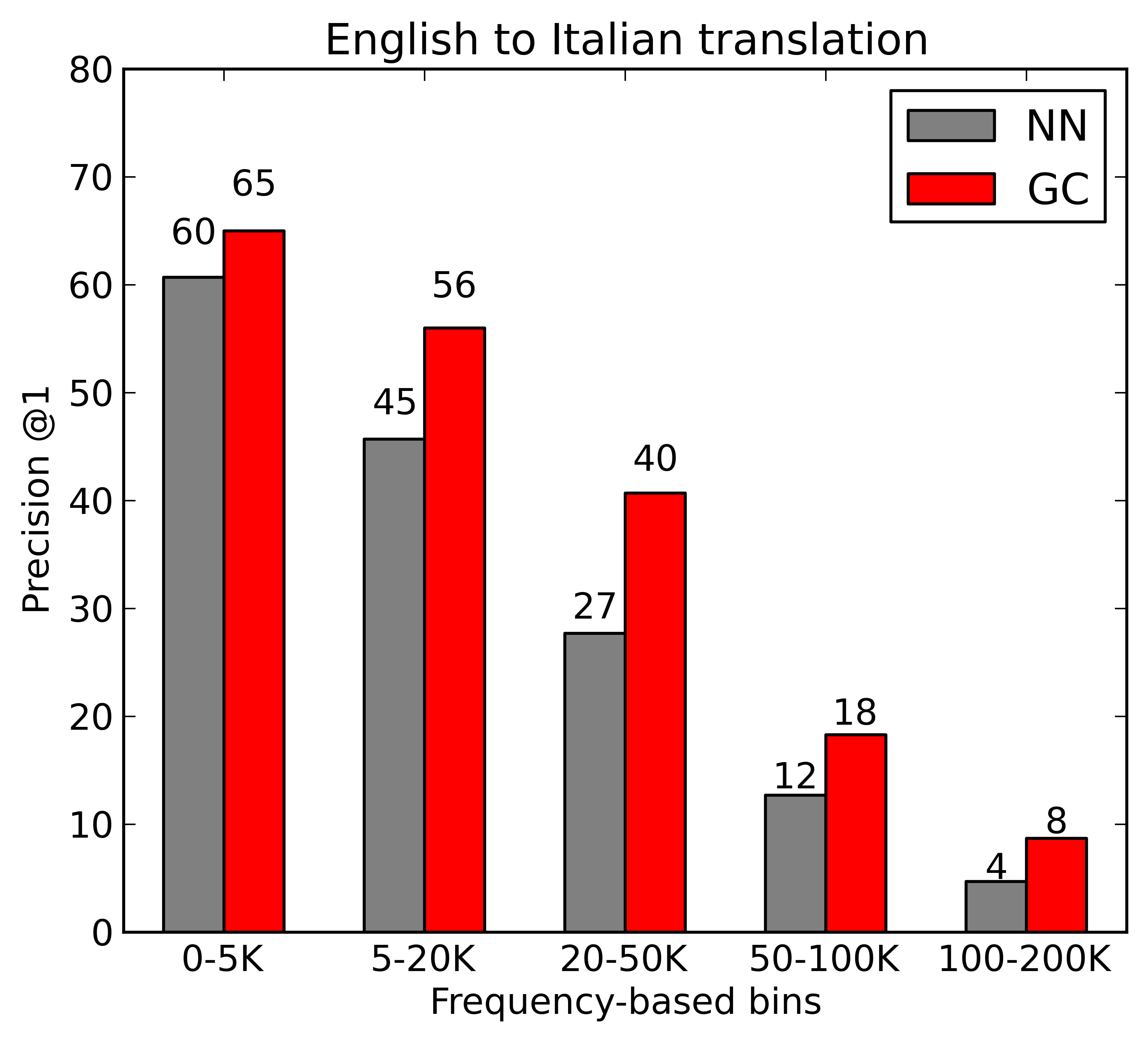}
\end{minipage}
\caption{Percentage accuracy scores for En$\to$It translation. Left:
  No regularization and Generalized Cross-validation (GCV) regression
  varying the training size. Right: Results split by frequency bins,
  5K training and no regularization.}
\label{experiments-lang-1}
\end{figure}

\paragraph{Results} Results are given in Figure
\ref{experiments-lang-1}. We report results without regularization as
well as with the regularization parameter $\lambda$ estimated by
generalized cross-validation (GCV) \citep[p.~244]{Hastie:etal:2009}. 
Both corrected methods achieve significant improvements over standard
$\reg{NN}$, ranging from 7\% to 14\%. For the standard method, the
performance decreases as the training data size increases beyond 5K,
probably due to the noise added by lower-frequency words.  The
corrected measures are robust against this effect: adding more
training data does not help, but it does not harm them
either. Regularization does not improve, and actually hampers the
standard method, whereas it benefits the corrected measures when using
a small amount of training data (1K), and does not affect performance
otherwise. The results by frequency bin show that most of the
improvements are brought about for the all-important medium- and
low-frequency words.  Although not directly comparable, the absolute
numbers we obtain are in the range of those reported by
\cite{mikolov2013exploiting}, whose test data correspond, in terms of
frequency, to those in our first 2 bins. Furthermore, we observe,
similarly to them, that the accuracy scores underestimate the actual
performance, as many translations are in fact correct but not present
in our gold dictionary.

\begin{figure}
\begin{center}
\begin{minipage}[c]{0.49\linewidth}
\centering
\begin{tabular}{|l|l|}
\hline
Hub & $\reg{N}_{20}$ \\
\hline
	blockmonthoff & 40\\
	04.02.05      & 26\\
	communautés   & 26\\
	limassol      & 25\\
	andè          & 23\\
	ampelia       & 23\\
	11/09/2002    & 20\\
	cgsi          & 19\\
	100.0         & 18\\
	cingevano     & 18\\
\hline
\end{tabular}
\end{minipage}
\begin{minipage}[c]{0.49\linewidth}
\centering
\includegraphics[scale=0.381]{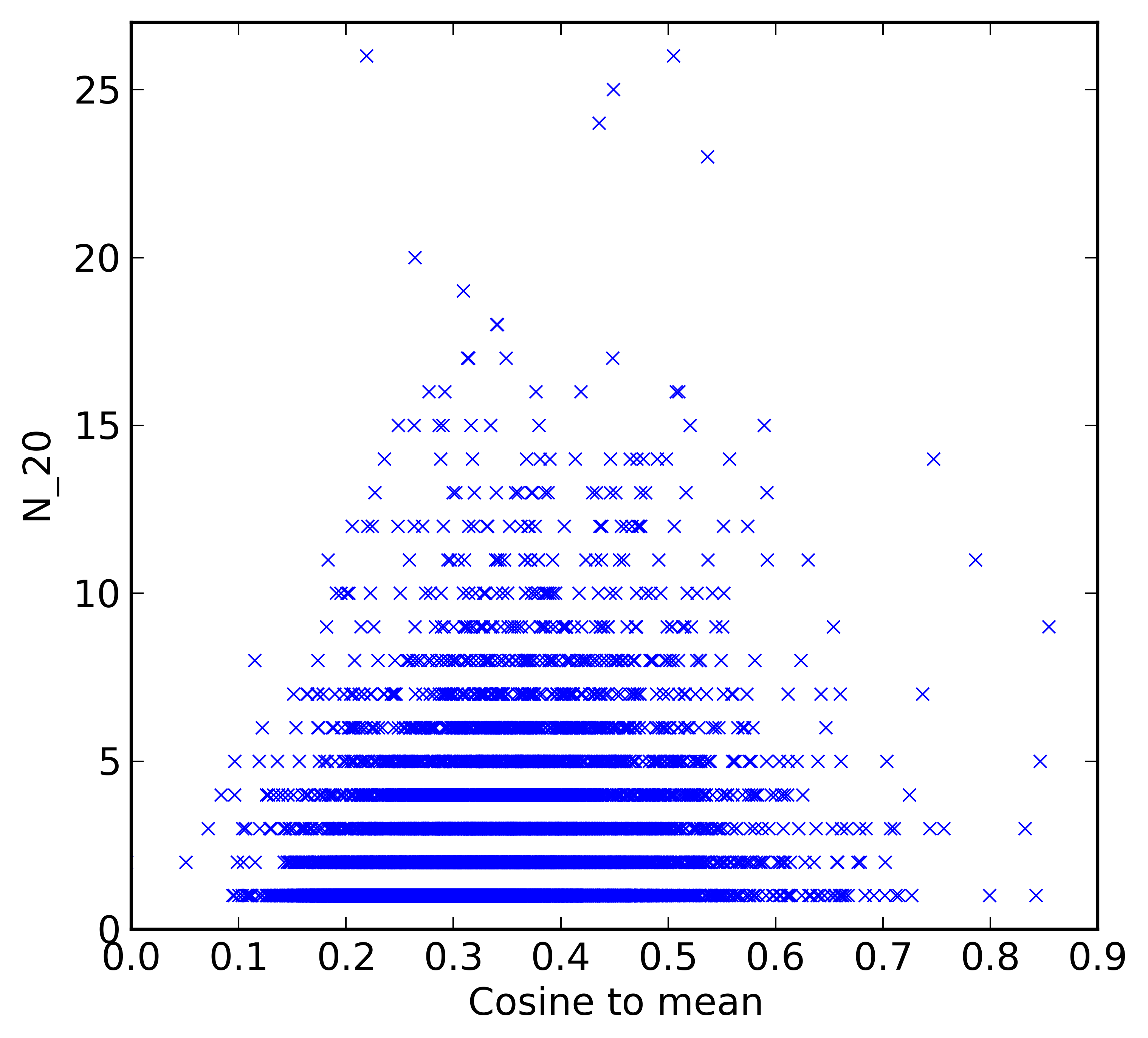}
\end{minipage}
\end{center}
\caption{En$\to$It translation. Left: Top 10 hubs and their
  $\reg{N}_{20,Tst}$ scores. Right: $\reg{N}_{20}$ plotted against
  cosine similarities to test-set mean vector. $\reg{N}_{20}$ values
  correlate significantly with cosines (Spearman $\rho=0.30$,
  $p=1.0e-300$).}
\label{experiments-lang-3}
\end{figure}

The elements with the largest hub score are shown in Figure
\ref{experiments-lang-3} (left). As can be seen, they tend to be
``garbage'' low-frequency words. However, in any realistic setting
such low-frequency terms should not be filtered out, as good
translations might also have low frequency. As pointed out by
\cite{Radovanovic:etal:2010b}, hubness correlates with proximity to
the test-set mean vector (the average of all test vectors). Hubness
level is plotted against cosine-to-mean in Figure
\ref{experiments-lang-3} (right).

\begin{table}
\begin{small}
\centering
\begin{tabular}{|l|l@{}c@{}l@{}|}
\hline
 &  Translation & $N_{20}(\reg{Hub})$& $\quad x|\reg{Hub}=\reg{NN_1}(x)$\\
\hline
almighty$\to$onnipotente & NN:dio         & 38 & righteousness,almighty,jehovah,incarnate,god...\\
Hub: dio (god)		   				 & GC: onnipotente & 20 & god\\
\hline
killers$\to$killer & NN: violentatori      & 64 & killers,anders,rapists,abusers,ragnar\\
Hub: violentatori (rapists)	& GC: killer & 22 & rapists\\
\hline
backwardness$\to$arretratezza & NN: 11/09/2002   & 110 & backwardness,progressivism,orthodoxies...\\
Hub: 11/09/2002 	& GC: arretratezza & 24 & orthodoxies,kumaratunga\\
\hline
\end{tabular}
\end{small}
\caption{En$\to$It translation. Examples of $\reg{GC}$ ``correcting'' $\reg{NN}$. The wrong $\reg{NN}$ translation is always a polluting hub. For these  $\reg{NN}$ hubs, we also report their hubness scores before and after correction ($\reg{N}_{20}$, over the whole pivot set) and examples of words they are nearest neighbour of ($x|\reg{Hub}=\reg{NN_1}(x)$) for $\reg{NN}$ and  $\reg{GC}$.}
\label{experiments-lang-2}
\end{table}

Table \ref{experiments-lang-2} presents some cases where wrong
translation are ``corrected'' by the $\reg{GC}$ measure. The latter
consistently pushes high-hubness elements down the neighbour lists.
For example, \emph{11/09/2002}, that was originally returned as the
translation of \emph{backwardness}, can be found in the $\reg{N}_{20}$
list of 110 English words. With the corrected method, the right
translation, \emph{arretratezza}, is obtained. \emph{11/09/2002} is
returned as the translation, this time, of only two other English
pivot words: \emph{orthodoxies} and \emph{kumaratunga}. The hubs we
correct for are not only garbage ones, such as \emph{11/09/2002}, but
also more standard words such as \emph{dio} (\emph{god}) or
\emph{violentatori} (\emph{rapists}), also shown in Table
\ref{experiments-lang-2}.\footnote{Prompted by a reviewer, we also performed preliminary experiments with a margin-based ranking objective similar to the one in WSABIE~\cite{Weston:etal:2011} and DeViSE~\cite{Frome:etal:2013} which is typically reported to outperform the l2 objective in Equation \ref{eq:obj1} (\cite{Socher:etal:2014}). Given a pair of training items ($\mathbf{x}_i, \mathbf{y}_i$) and the corresponding prediction $\mathbf{\hat{y}}_i=\mat{W}\mathbf{x}_i$, the error is given by: $\sum_{j=1, j \neq i}^k \max\{0,\gamma+dist(\mathbf{\hat{y}}_i, \mathbf{y}_i)-dist(\mathbf{\hat{y}}_i,\mathbf{y}_{j})\} \label{eq:obj2}$, where $dist$ is a distance measure, which we take to be inverse cosine, and $\gamma$ and $k$ are the margin and the number of negative examples, respectively. We tune $\gamma$-the margin and $k$-the number of negative samples on a held-out set containing 25\% of the training data. We estimate $\mat{W}$ using stochastic gradient descent where per-parameter learning rates are tuned with Adagrad~\cite{Duchi:etal:2011}. 
Results on the En$\to$It task are at 38.4 (NN) and further improved to 40.6 (GC retrieval), confirming that GC is not limited to least-squares error estimation settings.}

\subsection{Zero-shot image labeling and retrieving}

In this section we test our proposed method in a cross-modal setting,
mapping images to word labels and vice-versa.
\paragraph{Experimental setting}
We use the data set of \cite{lazaridou2014combining} containing 5,000
word labels, each associated to 100 ImageNet pictures
\citep{Deng:etal:2009}. Word representations are extracted from
Wikipedia with word2vec in skip-gram mode. Images are represented by
4096-dimensional vectors extracted using the Caffe toolkit
\citep{Jia:2014:CCA:2647868.2654889} together with the pre-trained convolutional
neural network of \cite{Krizhevsky:etal:2012}. We use a random 4/1
train/test split.
\paragraph{Results}

We consider both the usual image labeling setting
(Vision$\to$Language) and the image retrieval setting
(Language$\to$Vision). For the Vision$\to$Language task, we use as
pivot set the 100K test images (1,000 labels x 100 images/label) and
an additional randomly chosen 100K images. The search space is the
entire label set of 5,000 words. For Language$\to$Vision, we use as
pivot set the entire word list (5,000) and the target set is the
entire set of images (500,000). The objects depicted in the images
form a set of 5,000 distinct elements, therefore, for the word
\emph{cat}, for example, returning any of the 100 cat images is
correct. Chance accuracy in both settings is thus at 1/5,000. 
Table \ref{experiments-vis-1} reports accuracy scores.\footnote{The
  non-regularized objective led to very low results in both directions
  and for all methods, and we omit these results.}  We observe that,
differently from the translation case, correcting by normalizing the
cosine scores of the elements in the target domain
($\reg{NN}_{\reg{nrm}}$) leads to poorer results than no
correction. On the other hand, the $\reg{GC}$ method is consistent
across domains, and it improves significantly on the standard
$\reg{NN}$ method in both settings.  Note that, while there are
differences between the setups, \cite{Frome:etal:2013} report accuracy
results below 1\% in all their zero-shot experiments, including those
with chance levels comparable to ours.

\begin{table}
\centering
\begin{tabular}{c|c|ccc}
 Train & Chance  & \multicolumn{3}{|c}{GCV}\\ 
\hline
						&& $\reg{NN}$ & $\reg{NN}_{\reg{nrm}}$& $\reg{GC}$\\
Vis$\to$Lang & 0.02 & 1.0 & 0.8 & 1.5\\
Lang$\to$Vis & 0.02 & 0.6 & 1.4 & 2.2\\
\hline
\end{tabular}
\caption{Percentage label and image retrieval accuracy.}
\label{experiments-vis-1}
\end{table}

\begin{figure}
\centering
\subfloat[Original]{\includegraphics[scale=0.35]{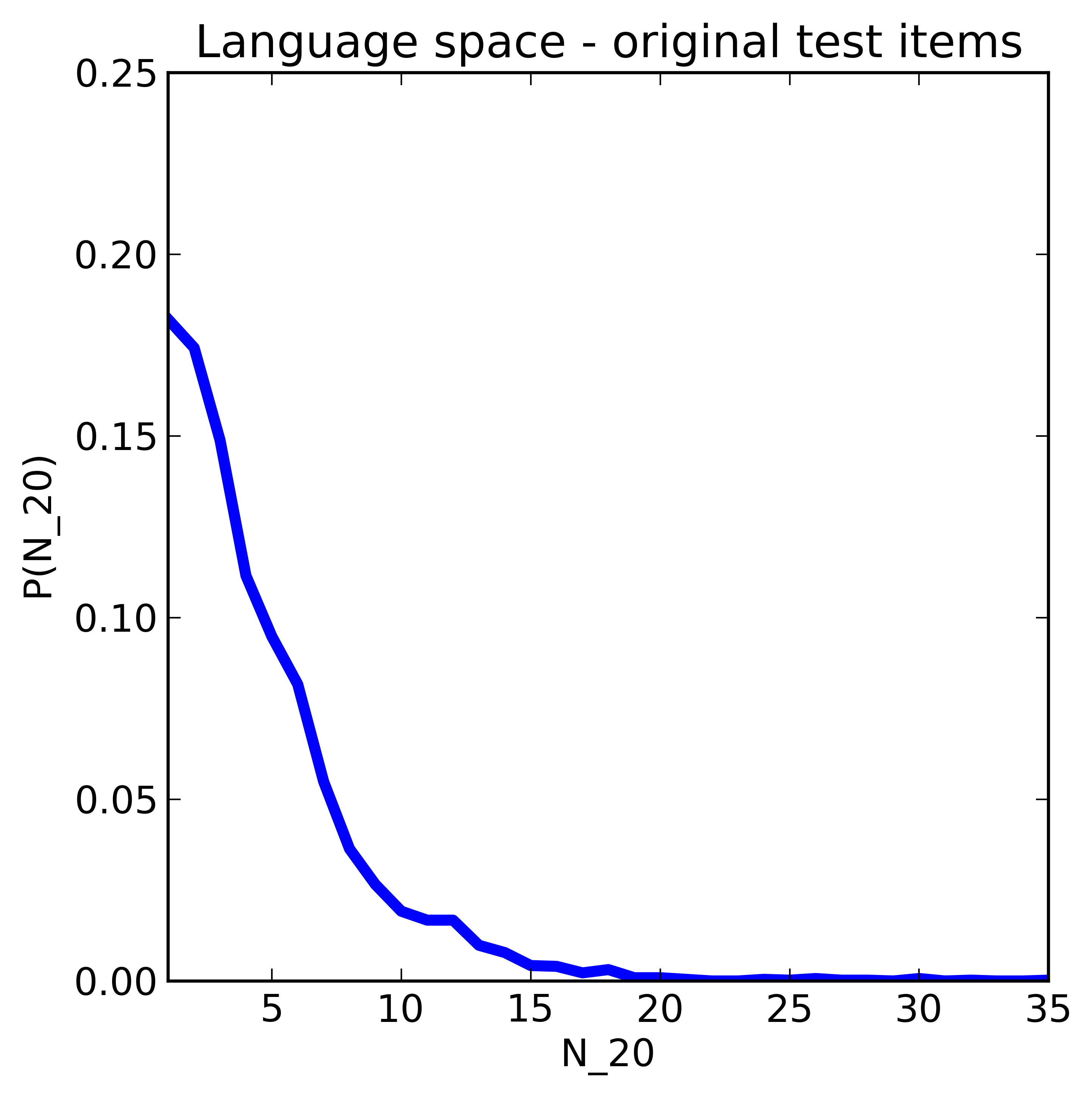}} 
\subfloat[Mapped]{\includegraphics[scale=0.35]{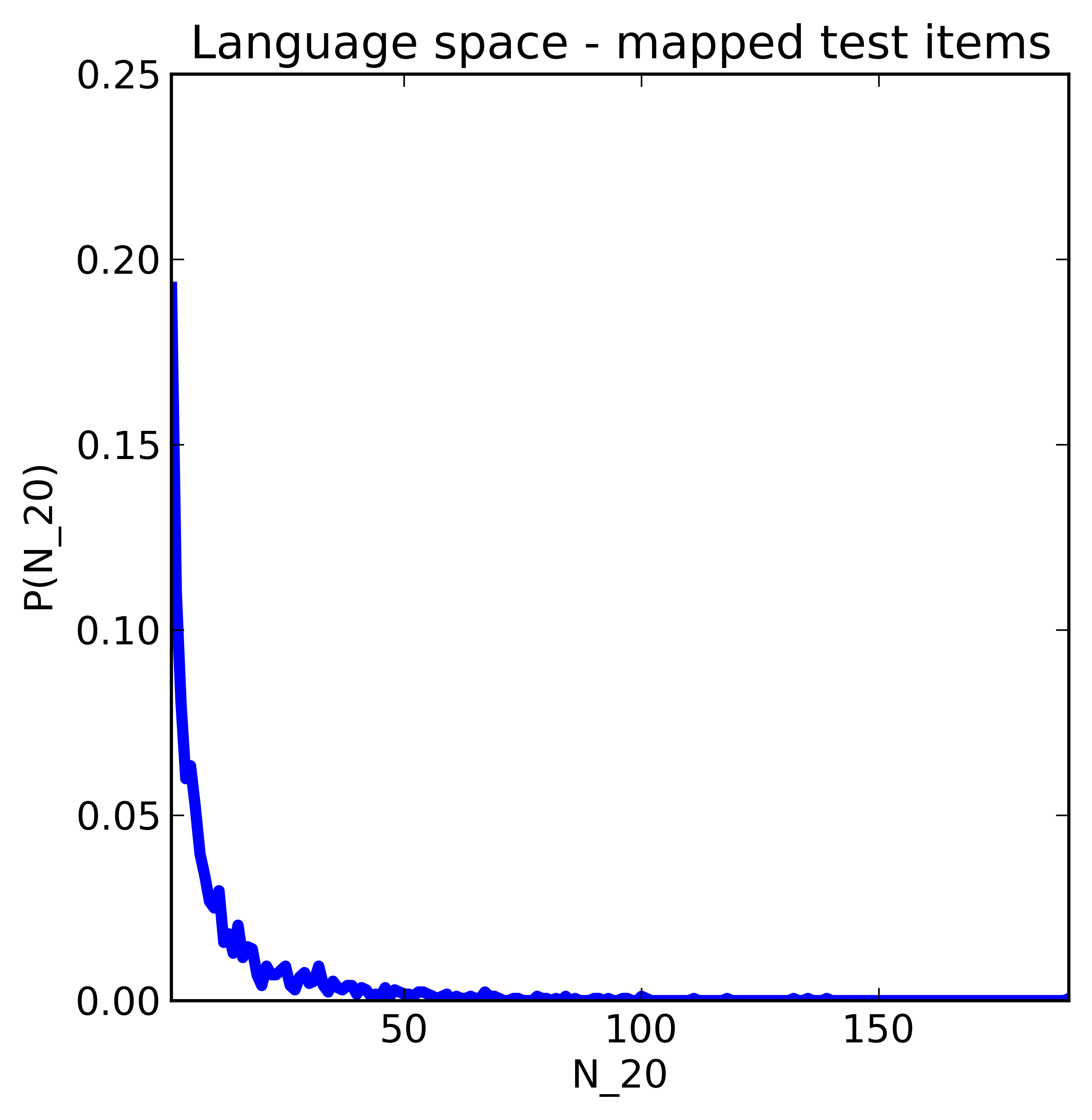}}
\subfloat[Corrected]{\includegraphics[scale=0.35]{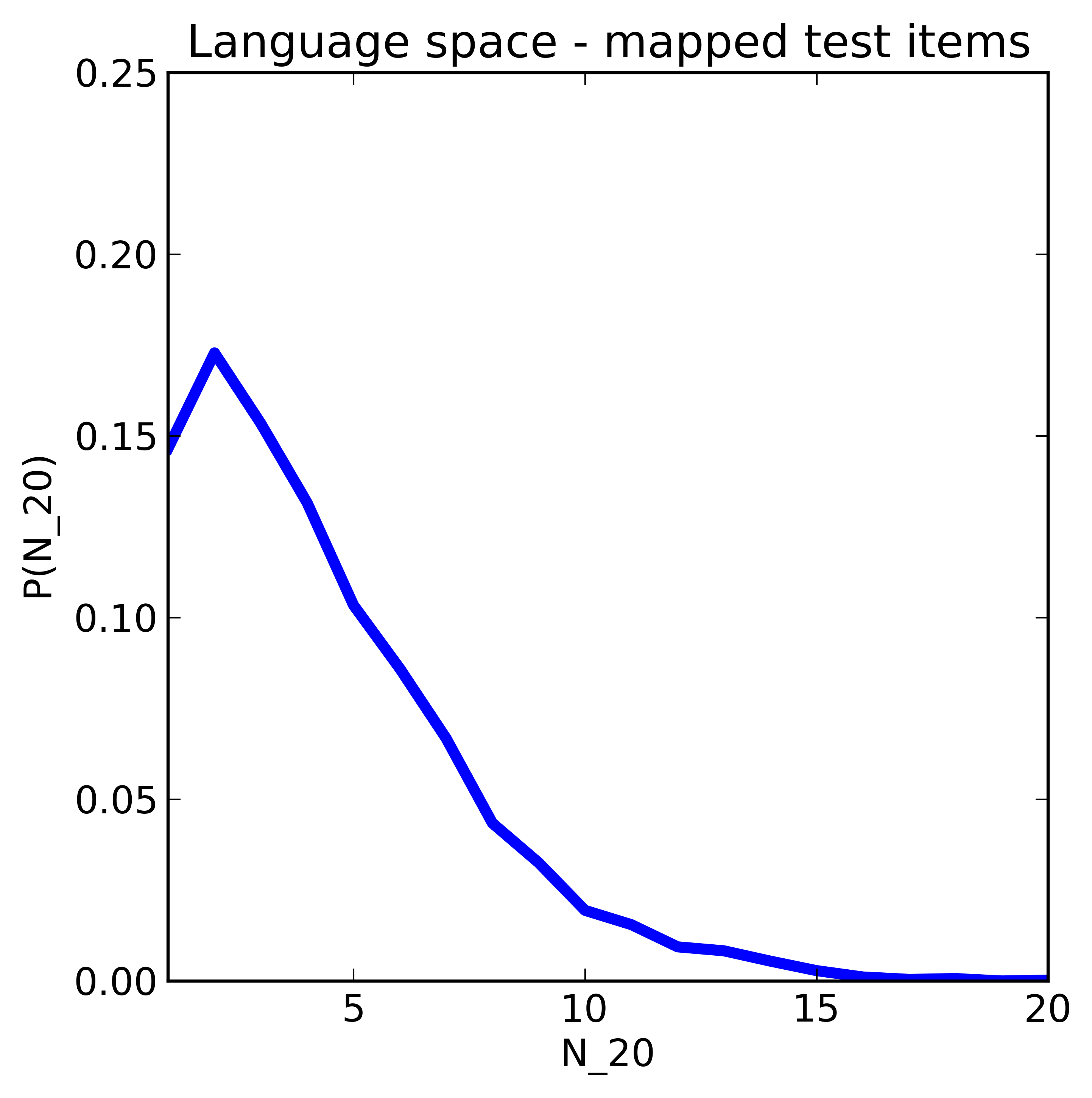}}
\caption{Vision$\to$Language. Distribution of $\reg{N}_{20}$ values of
  target space elements ($\reg{N}_{20,Ts}(y)$ for $y\in T$). Pivot
  vectors of the gold word labels of test elements in target (word)
  space (\emph{original}) vs.~corresponding mapped vectors
  (\emph{mapped}) vs.~corrected mapped vectors ($\reg{GC}$)
  (\emph{corrected}). $\reg{N}_{20}$ values increase from original to
  mapped sets (max 35 vs.~190), but they drop when using $\reg{GC}$
  correction (max 20).}
  \label{hubness vision full space query}
\label{experiments-vision-3}
\end{figure}

In order to investigate the hubness of the corrected solution, we plot
similar figures as in Section \ref{sec:hubness-problem}, computing the
$\reg{N}_{20}$ distribution of the target space elements w.r.t~the
pivots in the test set.\footnote{In order to facilitate these
  computations, we use ``aggregated'' visual vectors corresponding to
  each word label (e.g., we obtain a single \emph{cat} vector in image
  space by averaging the vectors of 100 \emph{cat} pictures).} 
Figure \ref{experiments-vision-3} shows this distribution 1) for the
vectors of the gold word labels in language space, 2) the
corresponding Vision$\to$Language mapped test vectors, as well as 3)
$N_{20}$ values computed using $\reg{GC}$ correction.\footnote{We
  abuse notation here, as $N_{20}$ is defined as in Equation
  \ref{hub-def} for 1) and 2) and as $|\{y \in \reg{GC}_k(x,T)|x \in
  P'\}|$ for 3).} Similarly to the translation case, the maximum
hubness values increase significantly from the original target space
vectors to the mapped items. When adjusting the rank with the
$\reg{GC}$ method, hubness decreases to a level that is now below
that of the original items. We observe the same trend in the
Language$\to$Vision direction (as well as in the translation
experiments in the previous section), the specifics of which we
however leave out for brevity.

\section{conclusion}

In this paper we have shown that the basic setup in zero-shot
experiments (use multivariate linear regression with a regularized
least-squares error objective to learn a mapping across
representational vectors spaces) is negatively affected by strong
hubness effects. We proposed a simple way to correct for this by
replacing the traditional nearest neighbour queries with globally
adjusted ones. The method only requires the availability of more,
unlabeled source space data, in addition to the test instances. While
more advanced ways for learning the mapping could be employed (e.g.,
incorporating hubness avoidance strategies into non-linear functions
or different learning objectives), we have shown that consistent
improvements can be obtained, in very different domains, already with
our query-time correction of the basic learning setup, which is a
popular and attractive one, given its simplicity, generality and high
performance. In future work we plan to investigate whether the hubness
effect carries through to other setups: For example, to what extent
different kinds of word representations and other learning objectives are affected by it. This empirical work should
pose a solid basis for a better theoretical understanding of the
causes of hubness increase in cross-space mapping.

\section{acknowledgments}
This work was supported by the ERC 2011 Starting Independent Research Grant n. 283554 (COMPOSES).

\bibliography{angeliki,marco,georgiana}

\begin{thebibliography}{29}
\providecommand{\natexlab}[1]{#1}
\providecommand{\url}[1]{\texttt{#1}}
\expandafter\ifx\csname urlstyle\endcsname\relax
  \providecommand{\doi}[1]{doi: #1}\else
  \providecommand{\doi}{doi: \begingroup \urlstyle{rm}\Url}\fi

\bibitem[Baroni et~al.(2014)Baroni, Dinu, and Kruszewski]{Baroni:etal:2014}
Baroni, Marco, Dinu, Georgiana, and Kruszewski, Germ\'{a}n.
\newblock Don't count, predict! a systematic comparison of context-counting vs.
  context-predicting semantic vectors.
\newblock In \emph{Proceedings of ACL}, pp.\  238--247, Baltimore, MD, 2014.

\bibitem[Clark(2015)]{Clark:2012b}
Clark, Stephen.
\newblock Vector space models of lexical meaning.
\newblock In Lappin, Shalom and Fox, Chris (eds.), \emph{Handbook of
  Contemporary Semantics, 2nd ed.} Blackwell, Malden, MA, 2015.
\newblock {I}n press;
  \url{http://www.cl.cam.ac.uk/~sc609/pubs/sem_handbook.pdf}.

\bibitem[Collobert et~al.(2011)Collobert, Weston, Bottou, Karlen, Kavukcuoglu,
  and Kuksa]{Collobert:etal:2011}
Collobert, Ronan, Weston, Jason, Bottou, L{\'e}on, Karlen, Michael,
  Kavukcuoglu, Koray, and Kuksa, Pavel.
\newblock Natural language processing (almost) from scratch.
\newblock \emph{Journal of Machine Learning Research}, 12:\penalty0 2493--2537,
  2011.

\bibitem[Deng et~al.(2009)Deng, Dong, Socher, Li, and Fei-Fei]{Deng:etal:2009}
Deng, Jia, Dong, Wei, Socher, Richard, Li, Lia-Ji, and Fei-Fei, Li.
\newblock Imagenet: A large-scale hierarchical image database.
\newblock In \emph{Proceedings of CVPR}, pp.\  248--255, Miami Beach, FL, 2009.

\bibitem[Dinu \& Baroni(2014)Dinu and Baroni]{Dinu:Baroni:2014}
Dinu, Georgiana and Baroni, Marco.
\newblock How to make words with vectors: Phrase generation in distributional
  semantics.
\newblock In \emph{Proceedings of ACL}, pp.\  624--633, Baltimore, MD, 2014.

\bibitem[Duchi et~al.(2011)Duchi, Hazan, and Singer]{Duchi:etal:2011}
Duchi, John, Hazan, Elad, and Singer, Yoram.
\newblock Adaptive subgradient methods for online learning and stochastic
  optimization.
\newblock \emph{The Journal of Machine Learning Research}, 12:\penalty0
  2121--2159, 2011.

\bibitem[Frome et~al.(2013)Frome, Corrado, Shlens, Bengio, Dean, Ranzato, and
  Mikolov]{Frome:etal:2013}
Frome, Andrea, Corrado, Greg, Shlens, Jon, Bengio, Samy, Dean, Jeff, Ranzato,
  {Marc'Aurelio}, and Mikolov, Tomas.
\newblock {DeViSE}: A deep visual-semantic embedding model.
\newblock In \emph{Proceedings of NIPS}, pp.\  2121--2129, Lake Tahoe, NV,
  2013.

\bibitem[Haghighi et~al.(2008)Haghighi, Liang, Berg-Kirkpatrick, and
  Klein]{HaghighiLBK08}
Haghighi, Aria, Liang, Percy, Berg-Kirkpatrick, Taylor, and Klein, Dan.
\newblock Learning bilingual lexicons from monolingual corpora.
\newblock In \emph{Proceedings of ACL}, pp.\  771--779, Columbus, OH, USA, June
  2008.
\newblock URL \url{http://www.aclweb.org/anthology/P/P08/P08-1088}.

\bibitem[Hastie et~al.(2009)Hastie, Tibshirani, and Friedman]{Hastie:etal:2009}
Hastie, Trevor, Tibshirani, Robert, and Friedman, Jerome.
\newblock \emph{The Elements of Statistical Learning, 2nd edition}.
\newblock Springer, New York, 2009.

\bibitem[Jia et~al.(2014)Jia, Shelhamer, Donahue, Karayev, Long, Girshick,
  Guadarrama, and Darrell]{Jia:2014:CCA:2647868.2654889}
Jia, Yangqing, Shelhamer, Evan, Donahue, Jeff, Karayev, Sergey, Long, Jonathan,
  Girshick, Ross, Guadarrama, Sergio, and Darrell, Trevor.
\newblock Caffe: Convolutional architecture for fast feature embedding.
\newblock In \emph{Proceedings of the ACM International Conference on
  Multimedia}, MM '14, pp.\  675--678, New York, NY, USA, 2014. ACM.
\newblock ISBN 978-1-4503-3063-3.
\newblock \doi{10.1145/2647868.2654889}.
\newblock URL \url{http://doi.acm.org/10.1145/2647868.2654889}.

\bibitem[Klementiev et~al.(2012)Klementiev, Irvine, Callison-Burch, and
  Yarowsky]{Klementiev:2012}
Klementiev, Alexandre, Irvine, Ann, Callison-Burch, Chris, and Yarowsky, David.
\newblock Toward statistical machine translation without parallel corpora.
\newblock In \emph{Proceedings of EACL}, pp.\  130--140, Avignon, France, 2012.
\newblock ISBN 978-1-937284-19-0.
\newblock URL \url{http://dl.acm.org/citation.cfm?id=2380816.2380835}.

\bibitem[Koehn \& Knight(2002)Koehn and Knight]{Koehn02learninga}
Koehn, Philipp and Knight, Kevin.
\newblock Learning a translation lexicon from monolingual corpora.
\newblock In \emph{In Proceedings of ACL Workshop on Unsupervised Lexical
  Acquisition}, pp.\  9--16, Philadelphia, PA, USA, 2002.

\bibitem[Krizhevsky et~al.(2012)Krizhevsky, Sutskever, and
  Hinton]{Krizhevsky:etal:2012}
Krizhevsky, Alex, Sutskever, Ilya, and Hinton, Geoffrey.
\newblock {ImageNet} classification with deep convolutional neural networks.
\newblock In \emph{Proceedings of NIPS}, pp.\  1097--1105, Lake Tahoe, Nevada,
  2012.

\bibitem[Lazaridou et~al.(2014{\natexlab{a}})Lazaridou, Bruni, and
  Baroni]{Lazaridou:etal:2014}
Lazaridou, Angeliki, Bruni, Elia, and Baroni, Marco.
\newblock Is this a wampimuk? cross-modal mapping between distributional
  semantics and the visual world.
\newblock In \emph{Proceedings of ACL}, pp.\  1403--1414, Baltimore, MD,
  2014{\natexlab{a}}.

\bibitem[Lazaridou et~al.(2014{\natexlab{b}})Lazaridou, Pham, and
  Baroni]{lazaridou2014combining}
Lazaridou, Angeliki, Pham, The~Nghia, and Baroni, Marco.
\newblock Combining language and vision with a multimodal skip-gram model.
\newblock In \emph{NIPS workshop on Learning Semantics}, 2014{\natexlab{b}}.

\bibitem[Mikolov et~al.(2013{\natexlab{a}})Mikolov, Chen, Corrado, and
  Dean]{Mikolov:etal:2013b}
Mikolov, Tomas, Chen, Kai, Corrado, Greg, and Dean, Jeffrey.
\newblock Efficient estimation of word representations in vector space.
\newblock \url{http://arxiv.org/abs/1301.3781/}, 2013{\natexlab{a}}.

\bibitem[Mikolov et~al.(2013{\natexlab{b}})Mikolov, Le, and
  Sutskever]{mikolov2013exploiting}
Mikolov, Tomas, Le, Quoc, and Sutskever, Ilya.
\newblock Exploiting similarities among languages for {M}achine {T}ranslation.
\newblock \url{http://arxiv.org/abs/1309.4168}, 2013{\natexlab{b}}.

\bibitem[Mitchell et~al.(2008)Mitchell, Shinkareva, Carlson, Chang, Malave,
  Mason, and Just]{Mitchell:etal:2008}
Mitchell, Tom, Shinkareva, Svetlana, Carlson, Andrew, Chang, Kai-Min, Malave,
  Vincente, Mason, Robert, and Just, Marcel.
\newblock Predicting human brain activity associated with the meanings of
  nouns.
\newblock \emph{Science}, 320:\penalty0 1191--1195, 2008.

\bibitem[Palatucci et~al.(2009)Palatucci, Pomerleau, Hinton, and
  Mitchell]{Palatucci:etal:2009}
Palatucci, Mark, Pomerleau, Dean, Hinton, Geoffrey~E, and Mitchell, Tom~M.
\newblock Zero-shot learning with semantic output codes.
\newblock In \emph{Proceedings of NIPS}, pp.\  1410--1418, 2009.

\bibitem[Radovanovi\'{c} et~al.(2010{\natexlab{a}})Radovanovi\'{c}, Nanopoulos,
  and Ivanovi\'{c}]{Radovanovic:etal:2010a}
Radovanovi\'{c}, Milos, Nanopoulos, Alexandros, and Ivanovi\'{c}, Mirjana.
\newblock On the existence of obstinate results in vector space models.
\newblock In \emph{Proceedings of SIGIR}, pp.\  186--193, 2010{\natexlab{a}}.

\bibitem[Radovanovi\'{c} et~al.(2010{\natexlab{b}})Radovanovi\'{c}, Nanopoulos,
  and Ivanovi\'{c}]{Radovanovic:etal:2010b}
Radovanovi\'{c}, Milo\v{s}, Nanopoulos, Alexandros, and Ivanovi\'{c}, Mirjana.
\newblock Hubs in space: Popular nearest neighbors in high-dimensional data.
\newblock \emph{Journal of Machine Learning Research}, 11:\penalty0 2487--2531,
  2010{\natexlab{b}}.

\bibitem[Rapp(1999)]{Rapp99}
Rapp, Reinhard.
\newblock Automatic identification of word translations from unrelated english
  and german corpora.
\newblock In \emph{Proceedings of the 37th annual meeting of the Association
  for Computational Linguistics on Computational Linguistics}, ACL '99, pp.\
  519--526. Association for Computational Linguistics, 1999.

\bibitem[Socher et~al.(2013)Socher, Ganjoo, Manning, and Ng]{Socher:etal:2013a}
Socher, Richard, Ganjoo, Milind, Manning, Christopher, and Ng, Andrew.
\newblock Zero-shot learning through cross-modal transfer.
\newblock In \emph{Proceedings of NIPS}, pp.\  935--943, Lake Tahoe, NV, 2013.

\bibitem[Socher et~al.(2014)Socher, Le, Manning, and Ng]{Socher:etal:2014}
Socher, Richard, Le, Quoc, Manning, Christopher, and Ng, Andrew.
\newblock Grounded compositional semantics for finding and describing images
  with sentences.
\newblock \emph{Transactions of the Association for Computational Linguistics},
  2:\penalty0 207--218, 2014.

\bibitem[Tiedemann(2012)]{TIEDEMANN12.463}
Tiedemann, J\"org.
\newblock Parallel data, tools and interfaces in opus.
\newblock In \emph{Proceedings of the Eight International Conference on
  Language Resources and Evaluation (LREC'12)}, Istanbul, Turkey, 2012.

\bibitem[Tomasev et~al.(2011{\natexlab{a}})Tomasev, Brehar, Mladenic, and
  Nedevschi]{10.1109/ICCP.2011.6047899}
Tomasev, Nenad, Brehar, Raluca, Mladenic, Dunja, and Nedevschi, Sergiu.
\newblock The influence of hubness on nearest-neighbor methods in object
  recognition.
\newblock \emph{Intelligent Computer Communication and Processing (ICCP), 2011
  IEEE International Conference}, 2011{\natexlab{a}}.

\bibitem[Tomasev et~al.(2011{\natexlab{b}})Tomasev, Radovanovic, Mladenic, and
  Ivanovic]{conf/cikm/TomasevRMI11}
Tomasev, Nenad, Radovanovic, Milos, Mladenic, Dunja, and Ivanovic, Mirjana.
\newblock A probabilistic approach to nearest-neighbor classification: naive
  hubness bayesian knn.
\newblock In \emph{CIKM}, 2011{\natexlab{b}}.

\bibitem[Turney \& Pantel(2010)Turney and Pantel]{Turney:Pantel:2010}
Turney, Peter and Pantel, Patrick.
\newblock From frequency to meaning: Vector space models of semantics.
\newblock \emph{Journal of Artificial Intelligence Research}, 37:\penalty0
  141--188, 2010.

\bibitem[Weston et~al.(2011)Weston, Bengio, and Usunier]{Weston:etal:2011}
Weston, Jason, Bengio, Samy, and Usunier, Nicolas.
\newblock Wsabie: Scaling up to large vocabulary image annotation.
\newblock In \emph{Proceedings of IJCAI}, pp.\  2764--2770, 2011.

\end{thebibliography}
\bibliographystyle{iclr2015}

\end{document}